%% file: main.tex
\documentclass[10pt,twocolumn,letterpaper]{article}

\usepackage[pagenumbers]{wacv} 


\definecolor{wacvblue}{rgb}{0.21,0.49,0.74}
\usepackage[pagebackref,breaklinks,colorlinks,allcolors=wacvblue]{hyperref}

\usepackage{graphicx}
\usepackage{booktabs}
\usepackage{tikz}
\usepackage{pgfplots}
\usepackage{amssymb}
\pgfplotsset{compat=1.18}
\usepackage{multirow}
\usepackage{array}
\newcolumntype{C}[1]{>{\centering\arraybackslash}p{#1}}
\usepackage{algorithm}
\usepackage{algpseudocode}

\def\wacvPaperID{501} 
\def\confName{WACV}
\def\confYear{2027}

\title{ObjectGuidedMap: An Object-Aware Model for Robust Bird's Eye View Map Prediction in High-Occlusion Scenarios}

\author{Christian Bohn\\
University of Wuppertal\\
Germany\\
{\tt\small christian.bohn@uni-wuppertal.de}
\and
Harsh Yadav\\
University of Wuppertal\\
Germany\\
{\tt\small harsh.yadav@uni-wuppertal.de}
\and
Klaus Friedrichs\\
Aptiv\\
Germany\\
{\tt\small klaus.friedrichs@aptiv.com}
\and
Hasan Tercan\\
University of Wuppertal\\
Germany\\
{\tt\small tercan@uni-wuppertal.de}
\and
Tobias Meisen\\
University of Wuppertal\\
Germany\\
{\tt\small meisen@uni-wuppertal.de}
}

\begin{document}
\maketitle

\begin{abstract}
  Accurate bird’s eye view (BEV) map prediction is a critical component of autonomous driving systems, enabling downstream planning and control. Recent approaches such as Mask2Map achieve strong performance by predicting vectorized, instance-level map elements directly from multi-view images. However, existing BEV map prediction models operate largely independently of object-level scene understanding, despite the fact that traffic participants frequently occlude or implicitly encode information about road geometry. In this work, we propose a multitask architecture that jointly predicts BEV maps and 3D objects, explicitly enabling information exchange between the two tasks. Building upon Mask2Map, we introduce an object detection branch and incorporate cross-task feature fusion at both the BEV feature level and the instance-query level via deformable attention and cross-attention mechanisms. This design allows map instances to leverage object context while remaining robust to occlusions caused by traffic participants. To stabilize multitask learning, we employ epoch-dependent residual connections that gradually increase cross-task interaction during training. Experimental results on the nuScenes and Argoverse 2 datasets demonstrate that our approach consistently improves BEV map prediction accuracy, with particularly significant gains of $21\%$ $mAP$ in scenes containing heavy object occlusion. 
\end{abstract}

\section{Introduction}
\begin{figure}
    \centering
    \includegraphics[width=\columnwidth]{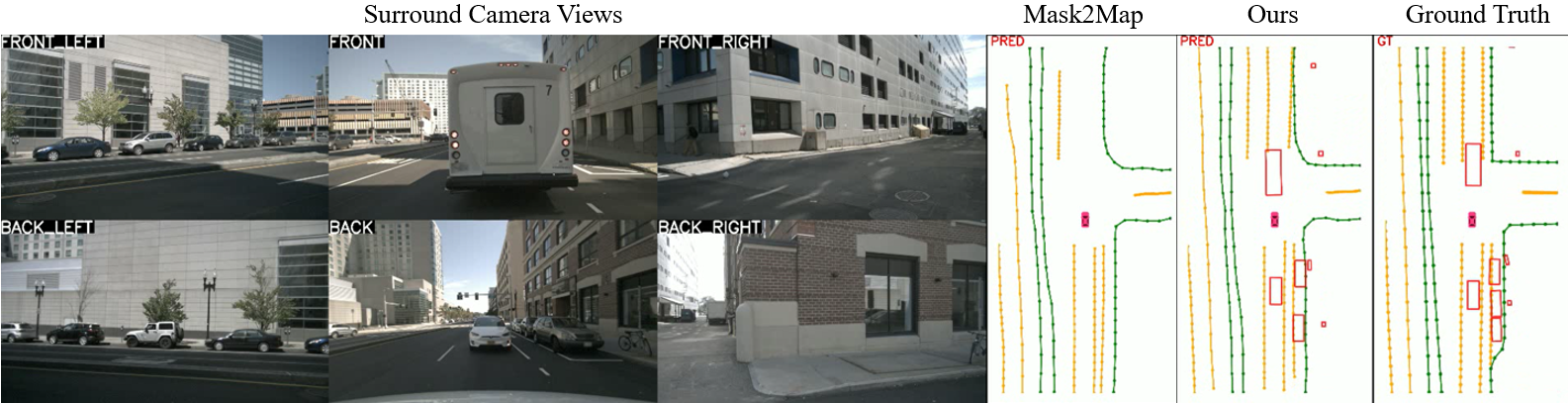}
    \caption{Comparison of predictions from our method to the Mask2Map \cite{mask2map} baseline.}
    \label{improvement_example_picture}
\end{figure}

The construction of Bird's Eye View (BEV) maps of the static environment around the ego-vehicle is a crucial part of the perception pipeline involved in autonomous driving \cite{HD_map_survey, vision_centric_bev_survey_paper}. While some early systems relied on pre-recorded High-Definition (HD) maps of their environment \cite{maps_for_autonomous_driving_survey, coarse-to-fine_localiaztion_with_hd_maps, vi-map}, this approach proved unreliable and difficult to scale. Modern systems rely on online BEV map construction \cite{maptr, maptrv2, mapvr, mask2map}, performed alongside object detection (OD) and the downstream trajectory prediction and planning tasks. The prediction of these maps usually relies on image input from an array of surround cameras, sometimes augmented with Radar or LiDAR data. Current BEV map predictors primarily reason geometrically but lack explicit relational conditioning on dynamic scene elements.\\ 
After some early works that modeled the task as a semantic segmentation problem \cite{lift_splat_shoot, bevformer, bevsegformer}, most modern methods replace the segmentation approach with \textit{vectorized map prediction}. Here, the map is represented as a set of \textit{map instances} which are individual elements making up the environment, such as lane lines or curbs. Each such map instance is represented by a sequence of points defining a polyline, alongside a class label. \\
Existing methods for constructing vectorized BEV maps, such as BeMapNet \cite{bemapnet}, MapTRv2 \cite{maptrv2}, and Mask2Map \cite{mask2map}, are already highly robust under ideal visibility conditions. However, poor weather, low lighting, and especially occlusions remain significant challenges that have not yet been resolved in the current literature. 
This is due to a key limitation in these existing approaches: They infer map geometry primarily from visible evidence in the image domain, without explicitly conditioning on the spatial configuration of surrounding traffic participants. In scenarios where map instances are occluded or their visibility is otherwise degraded, this lack of relational conditioning can hinder reliable reconstruction.
To systematically address this gap in the existing research, we introduce our novel \textit{ObjectGuidedMap} model. We hypothesize that enabling the model to take into account information about the objects present in the environment should make it more resilient against occlusions from these objects. Additionally, in poor-visibility scenarios, knowing the location of the surrounding objects should also provide a strong prior for the placement of certain map instances, \eg the position of a car parked on the side of the street should provide useful information for the position of the curb or a lane line. Hence, instead of performing the map prediction task in isolation, our approach introduces attention modules utilizing information from a parallel object detection branch in both the model's encoder and decoder. This enables our model to generate more robust BEV map predictions, as exemplified by the much improved prediction in \cref{improvement_example_picture}: Here, due to the large occlusion from the van in front of the ego-vehicle, the baseline model fails to predict the occluded lane lines, while our model's prediction is not impacted by the occlusion. \\

\begin{figure}
    \centering
    \includegraphics[width=\columnwidth]{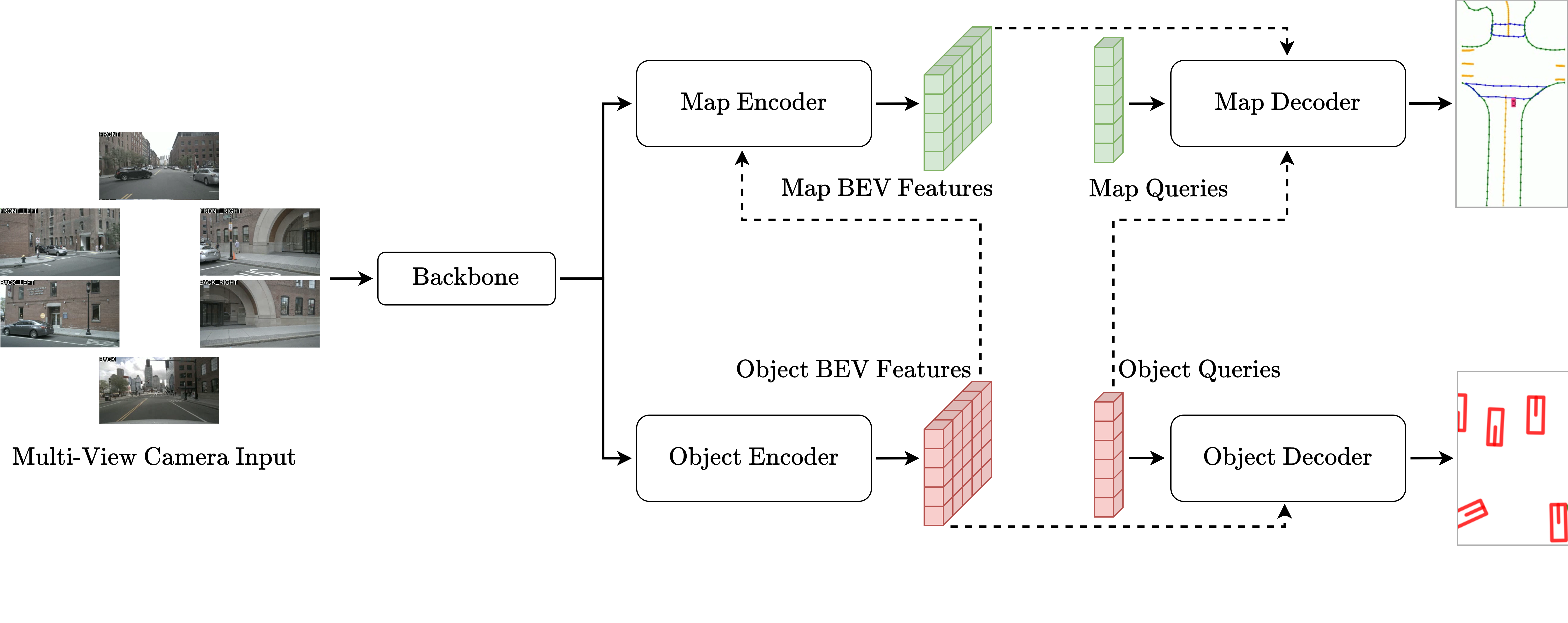}
    \caption{Overview of our multitask architecture: We propagate information from the object detection to the map prediction task in both the encoder and decoder of our model. Dashed lines indicate sampling operations.}
    \label{simple_arch_overview}
\end{figure}

To enable this awareness of the objects in the scene, we introduce two new attention modules into the well-established state-of-the-art Mask2Map \cite{mask2map} model. 
A simplified illustration of our resulting model architecture is shown in \cref{simple_arch_overview}. 

We evaluate our proposed ObjectGuidedMap method on the challenging and widely-used nuScenes \cite{nuscenes_paper} and Argoverse 2 \cite{av2_paper} datasets. 
Under our single-GPU training protocol, our approach improves the reproduced Mask2Map baseline by $4.99\%$, achieving a $mAP$ of $0.661$ on nuScenes.
For Argoverse 2, we demonstrate a similar effectiveness, with our method achieving a $7.57\%$ improvement relative to the baseline. 
Further investigating the source of these improvements, we show that this overall improvement over the full dataset is mainly due to high-occlusion or poor-visibility scenes in the dataset where our model outperforms the baseline by up to $21\%$ in the $mAP$ metric, while the performance improvement for scenarios with good visibility is negligible. This demonstrates the effectiveness of our method in scenes where awareness of the surrounding objects is crucial for precise map reconstruction. 

The key contributions of our work are: 
\begin{itemize}
    \item We propose a multitask architecture that jointly predicts objects and vectorized BEV maps. Unlike prior isolated map prediction models, our approach explicitly leverages object-level scene understanding as geometric and contextual cues to reconstruct complex map elements.
    \item We introduce cross-attention modules that enable information sharing between the object and map branches at both the encoder and decoder levels. As joint training can be unstable, we also introduce an effective epoch-dependent residual connection that gradually increases feature sharing during training. 
    \item We evaluate our approach on the nuScenes and Argoverse 2 datasets, demonstrating that it outperforms previous state-of-the-art methods for matched training conditions. Furthermore, our object-aware map predictions make the model robust to occluded areas, resulting in a 21\% boost in mAP for heavily occluded scenes. This confirms our hypothesis that awareness of surrounding objects improves resilience against occlusions.
\end{itemize}

\section{Related Work}
\textbf{Bird's Eye View Map Prediction.}
Bird's Eye View (BEV) map prediction is a core perception task for autonomous driving, aiming to infer a structured representation of the static driving environment from on-board sensors. 
Early learning-based approaches focused on predicting rasterized BEV semantic maps \cite{gosala22bev, lift_splat_shoot, cvt, bevformer, bevsegformer, petrv2, reiher2020sim2realdeeplearningapproach, roddick2020predictingsemanticmaprepresentations, cobevt, beverse}. 
While raster representations are convenient as they are straightforward to optimize with a simple pixel-wise cross-entropy loss, they inherently lack geometric precision, as uncertainty is reflected by predicting thicker or blurred map elements rather than committing to exact geometry. Moreover, rasterized maps use a uniform resolution across the entire BEV plane, leading to substantial redundancy by densely representing large regions devoid of map elements, and are therefore suboptimal for downstream planning and reasoning modules that require compact instance-level map representations \cite{Yadav_2025_CVPR, qcnet, scenetransformer, sept, vectornet, lanegcn}.
To address these limitations, recent works have shifted towards \emph{vectorized} and \emph{instance-level} BEV map representations. HDMapNet \cite{hdmapnet} bridges the gap between rasterized and vectorized map representations by post-processing its predicted semantic segmentation map and by adding an instance embedding into a vectorized BEV map. 
To overcome the inefficiencies associated with that post-processing approach, MapTR \cite{maptr} formulates map instances as polylines with a fixed number of points and predicts them using a transformer decoder with learnable instance queries inspired by the DETR \cite{detr} architecture. 
VectorMapNet \cite{vectormapnet} splits this approach into two separate modules: For each map instance, it first produces a set of rough keypoints that are subsequently further refined into a detailed polyline. With this refinement method, the authors achieved improved prediction quality compared to MapTR. 
MapVR \cite{mapvr} combines vectorized map prediction with a downstream differentiable rasterization module to provide supervision in both the vectorized and rasterized domains.
PivotNet \cite{pivotnet} replaces the common representation of a map instance with a more flexible but also more complex representation: Instead of predicting a fixed number of points for each instance, it predicts a sequence of pivots which are more densely spaced in high-curvature regions and sparser for straight parts of map elements. 
BeMapNet \cite{bemapnet} represents a map instance as piecewise B\'ezier Curve between a variable-length sequence of control points. Combined with a loss constraining the region mask produced from each predicted instance, this allows for a more precise map reconstruction compared to a polyline representation. 
MapTRv2 \cite{maptrv2} further refines the MapTR formulation by introducing auxiliary supervision and a permutation-invariant loss that accounts for all correct point orderings of a predicted map instance, resulting in improved training stability and consistently better performance across multiple benchmarks.
Mask2Map \cite{mask2map} advances vectorized BEV map prediction by introducing instance-level mask prediction. By first predicting a coarse spatial mask for each map instance and then refining the polyline prediction within the masked region, the Mask2Map model achieves superior geometric accuracy, particularly in cluttered urban scenes. Due to its state-of-the-art performance and architectural flexibility, Mask2Map serves as the baseline for our work.
Despite these advances, existing BEV map prediction methods model the task in isolation, without explicitly incorporating information about objects present in the scene.\\
\textbf{BEV Object Detection from Multi-View Images.} In parallel to BEV map prediction, substantial progress has been made in 3D object detection in BEV using multi-camera image inputs. Early approaches explored constructing BEV representations by lifting image features into 3D space using geometric projection or learned depth estimation, as exemplified by works such as Lift-Splat-Shoot \cite{lift_splat_shoot}, OFT \cite{oft}, and MV3D \cite{mv3d}. Building upon these foundations, BEVFormer \cite{bevformer} pioneered a transformer-based framework that aggregates multi-view image features into a grid of BEV queries via deformable attention \cite{deformable_detr}. In this BEV grid, each cell corresponds to a specific location in the scene, enabling flexible spatial reasoning and effective fusion across camera views. Some version of this setup pioneered by BEVFormer is still used in the latest object detection models. 
Subsequent works further extend the line of research on BEV-based 3D object detection from multi-view images. 
BEVDet \cite{bevdet} introduces a modular pipeline that efficiently constructs BEV features from multi-camera inputs, achieving a favorable accuracy--efficiency trade-off, while 
BEVDet4D \cite{bevdet4d} extends this design with temporal feature aggregation to better exploit motion cues across frames. 
PETR \cite{petr} encodes image features into a 3D meshgrid in world coordinates that interacts with learnable object queries in the decoder to generate object proposals and  
PETRv2 \cite{petrv2} builds upon this formulation by incorporating temporal modeling and additional map prediction tasks. 
BEVDepth \cite{bevdepth} focuses on improving geometric accuracy through explicit depth supervision to refine the BEV features, leading to more reliable predictions. 
BEVFormerV2 \cite{bevformerv2} adapts the original BEVFormer architecture by introducing perspective-view supervision to improve gradient flow to the backbone of the model, achieving faster convergence. 
PolarFormer \cite{polarformer} explores polar-coordinate representations to better align the BEV feature space with ego-centric geometry.
BEVNeXt \cite{bevnext} improves the precision of BEV grids by using both depth supervision and temporal aggregation. 
StreamPETR \cite{streampetr} efficiently propagates historical information across frames by retaining the object queries from previous video frames in a queue and attending to them, while 
RoPETR \cite{ropetr} further improves temporal reasoning by enhancing the positional embeddings of these object queries. 
SOLOFusion \cite{solofusion} exploits both short- and long-term temporal cues to improve detection robustness.
Collectively, these works demonstrate that BEV representations are highly effective for object-level scene understanding from monocular or multi-view image inputs.
However, these models typically optimize the task of detecting objects in isolation, without considering their interaction with map elements or the potential benefits of object-aware map reasoning.\\
\textbf{Multitask Learning in Autonomous Driving Perception.} Multitask learning has been extensively studied as a means to improve efficiency and representation sharing across related tasks \cite{mtl_survey, crawshaw2020multitasklearningdeepneural, ruder2017overviewmultitasklearningdeep, ubernet, misra2016crossstitchnetworksmultitasklearning}. Several works employ shared backbones with task-specific heads, demonstrating improved performance and often reduced computational cost compared to independent models \cite{mti-net, mult, demt}.
In the BEV domain, some approaches jointly predict multiple outputs, including occupancy grids, semantic maps, and object detections \cite{fiery, stp3, bevformer, beverse, m2bev, maskbev, petrv2}. 
In BEVFormer \cite{bevformer}, M2BEV \cite{m2bev}, and BEVerse \cite{beverse}, the different tasks share a common backbone, encoder, and BEV representation but are decoded independently. MaskBEV\cite{maskbev} and PETRv2\cite{petrv2} employ a shared decoder for all task queries, but the queries corresponding to different tasks do not explicitly interact in a separate cross-task attention module.
While these approaches all share a common latent BEV representation and employ task-specific query sets, the interaction between tasks remains largely implicit and is typically confined to early-stage feature sharing in the encoder. Moreover, because each of these methods formulate BEV map prediction as semantic segmentation rather than vectorized instance prediction, each map query corresponds to a patch in the spatial grid instead of a distinct map instance. This representation precludes a direct, instance-level interaction between an object and a map instance. Consequently, explicit mechanisms that facilitate \emph{instance-level interaction} across tasks, such as enabling map instances to attend to detected objects, remain largely unexplored. 
Recently, most multitask learning works for autonomous driving focus on end-to-end training with sensor data as inputs and and an ego-vehicle trajectory as output \cite{hu2023_uniad, jiang2026vadv2endtoendvectorizedautonomous, Weng_2024_CVPR, sun2025sparsedrive, ye2023fusionadmultimodalityfusionprediction, jia2025drivetransformer}. However, these papers still perform the perception tasks in isolation and the queries from map- and object prediction only interact later downstream in the planning and prediction modules. An exception is the DualAD model by Doll \etal\cite{doll2024dualaddisentanglingdynamicstatic} whose object queries attend to the map queries, but not vice versa. In general, these works tend to treat perception task modules as black boxes whose functionality is abstracted away. \\
\textbf{Context-Aware and Occlusion-Robust Map Prediction.}
Handling occlusions is a fundamental challenge in autonomous driving perception. Several works leverage temporal aggregation \cite{bevformer, fiery, yuan2023streammapnetstreamingmappingnetwork, lin2023sparse4dv2recurrenttemporal} or motion cues \cite{stp3} to infer occluded scene elements. 
Other approaches explore context-aware reasoning by incorporating semantics about lane topology into their map predictions \cite{topologic, topopoint}. Nevertheless, these methods do not explicitly exploit detected objects as contextual cues for BEV map prediction. 
Using object detection outputs to directly guide vectorized BEV map instance prediction still remains underexplored.

\section{Method}
\begin{figure*}
    \centering
    \includegraphics[width=0.9\linewidth]{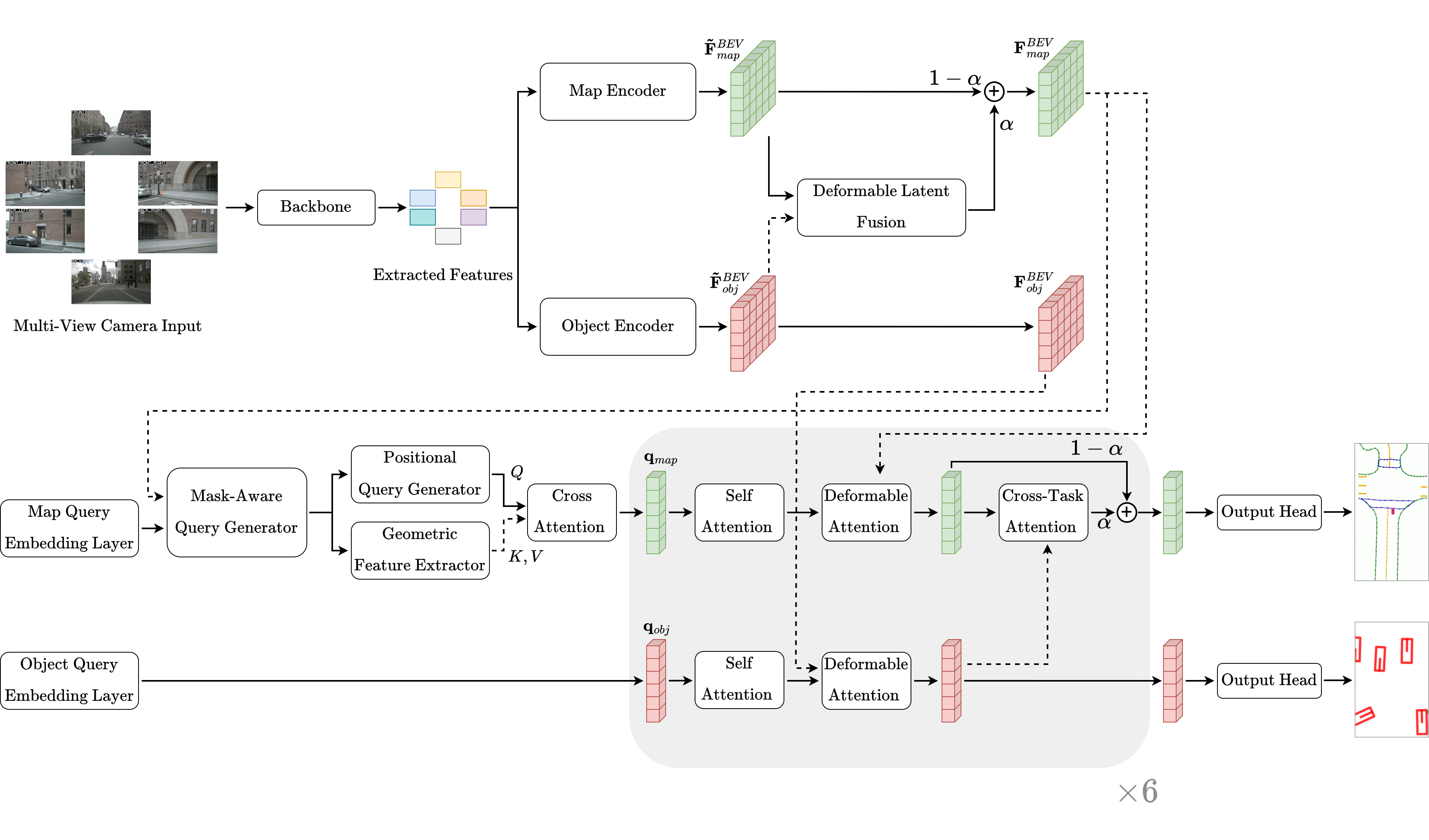}
    \caption{Overview of our Multitask Architecture: Dashed connecting lines indicate sampling operations.}
    \label{architecture_diagram}
\end{figure*}

An overview of our multitask architecture is shown in \cref{architecture_diagram}. After a shared feature extraction backbone both tasks are handled by their own task branches that are connected only by attention modules. The object detection branch (with its tensor depicted in red) follows the approach used in BEVFormer \cite{bevformer}, while the branch for the vectorized map prediction task (with its tensors colored green) builds on the architecture introduced in Mask2Map \cite{mask2map}. The top half of the figure depicts the encoder part of the network along with the backbone while the bottom shows the decoder. 

\subsection{Backbone} Our model takes as its only input a set of $c$ surround-view camera images of the traffic scene around the ego-vehicle. For each input image, the backbone network generates a feature map, resulting in a set of $c$ feature maps that can then effectively be used in the downstream modules of the network.

\subsection{Encoders} The encoders for both tasks transform the extracted feature maps from the backbone into grid-shaped BEV latent representations of size $[h_{map}, w_{map}]$ for the map prediction task and $[h_{obj}, w_{obj}]$ for the object detection task. Both lift the images features into 3D space using the intrinsics and extrinsics of the cameras. The Map Encoder uses the \textit{Lift-Splat-Shoot} operation \cite{lift_splat_shoot}, while the Object Encoder relies on Deformable Cross Attention \cite{deformable_detr} from the grid cells into the feature maps. 

\subsection{Deformable Latent Fusion} To facilitate information flow from the latent representation of the object detection task, $\mathbf{\tilde{F}}^{BEV}_{obj}$, to the map prediction BEV features, $\mathbf{\tilde{F}}^{BEV}_{map}$, we introduce the \textit{Deformable Latent Fusion (DLF)} module. It uses Deformable Attention \cite{deformable_detr} to allow the map prediction latent features to efficiently attend to the features of the object detection task, fusing object information into the map features. Due to the relatively large size of the BEV grids, computing traditional Multi-Head Attention \cite{attention_is_all_you_need} is not computationally feasible and the locality bias introduced by Deformable Attention is appropriate here. Since the aim in this work is improving BEV map prediction, we only fuse information unidirectionally between the tasks, leaving bidirectionality as future work. 
The final BEV grid-features, $\mathbf{F}^{BEV}_{map}$ and $\mathbf{F}^{BEV}_{obj}$ are computed as follows:
\begin{equation}
    \mathbf{F}^{BEV}_{obj} = \mathbf{\tilde{F}}^{BEV}_{obj}
\end{equation}
and 
\begin{equation}
    \mathbf{F}^{BEV}_{map} = (1-\alpha) \cdot \mathbf{\tilde{F}}^{BEV}_{map} + \alpha \cdot \text{DefAttn}(\text{q}=\mathbf{\tilde{F}}^{BEV}_{map}, \text{i}=\mathbf{\tilde{F}}^{BEV}_{obj})
\end{equation}
with the epoch-dependent attention factor $\alpha$ introduced below; q and i in DefAttn denote query and input. The reference point for any grid cell in the query tensor $\mathbf{\tilde{F}}^{BEV}_{map}$ is the position in the input tensor $\mathbf{\tilde{F}}^{BEV}_{obj}$ that corresponds to the same location in the ego-coordinate system.  

\subsection{Epoch-Dependent-Attention Factor}  In the computation of $\mathbf{F}^{BEV}_{map}$, directly adding the attention output to the residual connection resulted in suboptimal performance. 
We found that allowing the map prediction branch to learn in relative isolation during the early training stages led to more stable optimization. 
To achieve this, we introduce an epoch-dependent attention factor $\alpha$ that gradually increases the weight of the attention output relative to the residual connection over time. 
This progressively increases the influence of attention while maintaining stability in early training.

\subsection{Object Decoder} Each object candidate is represented by one object query, in total our model uses $n_{obj}$ such queries which together define the query tensor $\mathbf{q}_{obj}$. Originating from a learnable embedding, these queries are then iteratively refined over a total of 6 refinement layers. In each layer the set of queries first attends to itself and subsequently to the object BEV features $\mathbf{F}_{obj}^{BEV}$ using Deformable Cross Attention \cite{deformable_detr}. In the object output head, the final object queries are then processed into a confidence score, a predicted class, as well as the position, height, width, depth, and angle around the axis perpendicular to the ground plane. 

\subsection{Map Decoder} For the map prediction task, our network uses one query for each of the $p$ points along a polyline describing a map instance. For a set of $n'_{map}$ map instances, the model predicts a total of $n_{map} = p \cdot n'_{map}$ map queries. Originating from a learnable embedding layer, the $n_{map}$ queries $\mathbf{q}_{map}$ are first enriched with information from the BEV features $\mathbf{F}^{BEV}_{map}$. This process is performed in the \textit{Mask-Aware Query Generator}, \textit{Positional Query Generator}, \textit{Geometric Feature Extractor}, and a cross attention layer from the queries generated by the Positional Query Generator to the features computed by the Geometric Feature Extractor. Up to this point, the generation of the map instance queries is identical to the Mask2Map \cite{mask2map} model, for more detail on the aforementioned modules refer to that publication. Similarly to the object decoder, the map queries $\mathbf{q}_{map}$ are then passed into 6 refinement layers. Each layer consists of a self-attention module, Deformable Cross Attention into the BEV features $\mathbf{F}^{BEV}_{map}$ and Cross-Task-Attention to the set of object queries. 

\subsection{Cross-Task-Attention} One important addition to the model architecture is the \textit{Cross-Task-Attention} module that enables the map queries to incorporate information from the object queries. 
For object-awareness while decoding the map queries, this interaction module is crucial. 
It is implemented as Multi-Head Attention \cite{attention_is_all_you_need}, cross-attending from the map queries to the object queries. The resulting object information is added to the map queries with a relative weight of $\alpha$, the same epoch-dependent attention factor defined for the Deformable Latent Fusion above. 
The exact refinement process of the queries from both tasks in the model decoder is detailed in \cref{alg:query_refinement}. The function names MHA, DefAttn, FFN, and LN denote Multi-Head Attention, Deformable Attention, Feedforward Network, and Layer Normalization, respectively.
\begin{algorithm}
\caption{Query Refinement in Decoder}
\label{alg:query_refinement}
{\scriptsize
\begin{algorithmic}[1]

\Require $\mathbf{q}_{\text{obj}}^0, \mathbf{q}_{\text{map}}^0$
\For{$\ell \in \{1, \dots, 6\}$}
    \State $\mathbf{q}_{\text{obj}}^\ell \gets \mathbf{q}_{\text{obj}}^{\ell-1}$
    \State $\mathbf{q}_{\text{map}}^\ell \gets \mathbf{q}_{\text{map}}^{\ell-1}$
    \Statex 
    \State $\mathbf{q}_{\text{obj}}^\ell \gets \mathrm{LN^\ell_{obj, 1}}(\mathrm{MHA^\ell_{obj}}(\mathrm{q}=\mathbf{q}_{\text{obj}}^\ell, \mathrm{k}=\mathbf{q}_{\text{obj}}^\ell, \mathrm{v}=\mathbf{q}_{\text{obj}}^\ell))$
    \State $\mathbf{q}_{\text{obj}}^\ell \gets \mathrm{LN^\ell_{obj, 2}}(\mathrm{DefAttn^\ell_{obj}}(\mathrm{query} =         \mathbf{q}_{\text{obj}}^\ell,\,
        \mathrm{input} = \mathbf{F}_{\text{obj}}^{\mathrm{BEV}}))$
    \State $\mathbf{q}_{\text{obj}}^\ell \gets \mathrm{LN^\ell_{obj, 3}}(\mathrm{FFN^\ell_{obj}}(\mathbf{q}_{\text{obj}}^\ell))$
    \Statex 
    \State $\mathbf{q}_{\text{map}}^\ell \gets \mathrm{LN^\ell_{map, 1}}(\mathrm{MHA^\ell_{map, 1}}(\mathrm{q}=\mathbf{q}_{\text{map}}^\ell, \mathrm{k}=\mathbf{q}_{\text{map}}^\ell, \mathrm{v}=\mathbf{q}_{\text{map}}^\ell))$
    \State $\mathbf{q}_{\text{map}}^\ell \gets \mathrm{LN^\ell_{map, 2}}(\mathrm{DefAttn^\ell_{map}}(\mathrm{query} = \mathbf{q}_{\text{map}}^\ell,\,
        \mathrm{input} = \mathbf{F}_{\text{map}}^{\mathrm{BEV}}))$
    \State $\mathbf{q}_{\text{map}}^\ell \gets \mathrm{LN^\ell_{map, 3}}(\mathrm{FFN^\ell_{map, 1}}(\mathbf{q}_{\text{map}}^\ell))$
    \State $\tilde{\mathbf{q}}_{\text{map}}^\ell \gets \mathbf{q}_{\text{map}}^\ell$
    \State $\mathbf{q}_{\text{map}}^\ell \gets \mathrm{FFN^\ell_{map, 2}}(\mathrm{MHA^\ell_{map, 2}}(\mathrm{q}=\mathbf{q}_{\text{map}}^\ell, \mathrm{k}=\mathbf{q}_{\text{obj}}^\ell, \mathrm{v}=\mathbf{q}_{\text{obj}}^\ell))$
    \State $\mathbf{q}_{\text{map}}^\ell \gets \alpha\cdot\mathrm{LN^\ell_{map, 4}}(\mathbf{q}_{\text{map}}^\ell) + (1-\alpha)\cdot\tilde{\mathbf{q}}_{\text{map}}^\ell$
    
\EndFor \\
\Return $\{\mathbf{q}_{\text{obj}}^1, \dots, \mathbf{q}_\text{obj}^6, \mathbf{q}_{\text{map}}^1, \dots, \mathbf{q}_\text{map}^6\}$

\end{algorithmic}
}
\end{algorithm}
Identically to Mask2Map, during training, the resulting queries for each task and refinement layer $\ell$ are subsequently processed into map instances and object proposals and further propagated to the loss functions constraining the predictions. For inference, only the predictions from the last refinement layer are relevant.  

\subsection{Loss Functions} For the map prediction task, our method uses the same loss functions as Mask2Map, with the sum of all map prediction losses denoted as $\mathcal{L}_{map}$. We additionally add a Focal Loss\cite{FocalLoss8417976} and an L1 Loss term to constrain classification and regression of the bounding boxes predicted in each refinement layer. After scaling the sum of the object detection losses $\mathcal{L}_{obj}$ by a hyperparameter $\lambda$ to roughly match the magnitude of the map prediction losses, we obtain the final loss of the model as $\mathcal{L} = \mathcal{L}_{map} + \lambda\cdot\mathcal{L}_{obj}$

\section{Experiments}
\label{experiments}
We evaluate our method on the well-established nuScenes \cite{nuscenes_paper} and Argoverse2 \cite{av2_paper} datasets. The nuScenes dataset contains $1000$ scenes of $20$ seconds each, with keyframes annotated at $2Hz$. Roughly $28K$ of these keyframes are used for training and $6K$ for validation. For each keyframe, $6$ surround camera images are annotated with the bounding boxes and classes for the relevant objects in them as well as the map layout. Additionally, for evaluation only, we also use the LiDAR point information for each keyframe to identify the amount of occlusion in a scene. The Argoverse2 dataset consists of $1000$ scenarios of $15$ seconds each, annotated at $10Hz$. $700$ of the scenarios in the dataset are used for training and $150$ for validation, with the remainder reserved as the test set. For each frame, the training data contains $7$ surround-view images annotated with bounding boxes, their classes, and the map of the environment. 

For both datasets, we predict map instances over a range of $\pm15m$ laterally and $\pm30m$ longitudinally, following the practice in most prior works \cite{maptr, maptrv2, mask2map, vectormapnet, pivotnet}. For the object detection task, we define the perception range as $\pm51.2m$ in both directions, as in \cite{bevformer}. Keeping the practice established across most previous works on vectorized map prediction \cite{maptr, maptrv2, mask2map}, we evaluate three classes of map instances: Lane Lines, Crosswalks, and Curbs. Based on the Chamfer Distance between a ground-truth and a predicted map instance, we compute the Average Precision (AP) with matching thresholds of $\{0.5m, 1.0m, 1.5m\}$. We report the mean AP over these three thresholds for each of the three instance classes (Lane Line AP, Crosswalk AP, Curb AP), and all map instances (mAP). 

\begin{table*}[h!]
\centering
\caption{Results on the nuScenes validation set. The * indicates results reproduced using publicized code. For the reproduced results, we report the mean and standard deviation over four separate training runs. The percentages after the last row denote relative improvement relative to the Mask2Map baseline.}
\resizebox{0.8\linewidth}{!}{
\begin{tabular}{l||C{2.1cm}|C{2.1cm}|C{2.1cm}|C{2.1cm}||C{2.1cm}}
    \hline
     & \begin{tabular}{@{}c@{}}Lane Line\\AP $\uparrow$\end{tabular} &  \begin{tabular}{@{}c@{}}Crosswalk\\AP $\uparrow$\end{tabular} & \begin{tabular}{@{}c@{}}Curb\\AP $\uparrow$\end{tabular} &  mAP $\uparrow$ & NDS $\uparrow$ \\
    \hline
    MapTR \cite{maptr} & 0.463 & 0.515 &0.531 & 0.503 &$-$\\
    MapVR \cite{mapvr} & 0.477 & 0.544 & 0.514 & 0.512 &$-$\\
    PivotNet \cite{pivotnet} & 0.562 & 0.565 & 0.601 & 0.576 &$-$\\
    BeMapNet \cite{bemapnet} & 0.577 & 0.623 & 0.594 & 0.598 &$-$\\
    MapTRv2 \cite{maptrv2} & 0.598 & 0.624 & 0.624 & 0.615 &$-$\\
    Mask2Map* \cite{mask2map} & $0.627\pm0.018$ & $0.586\pm0.027$ & $0.674\pm0.019$ & $0.629\pm0.021$ &$-$ \\
    Ours & \textbf{0.655}$\pm0.009$ (+4.47\%) & \textbf{0.627}$\pm0.016$ (+6.97\%)& \textbf{0.699}$\pm0.001$ (+3.74\%) & \textbf{0.661}$\pm0.008$ (+4.99\%)&  0.309$\pm0.006$\\
    \hline
    \end{tabular}
    }
    \label{nuscenes_results_table}
\end{table*}

To maintain comparability of our model with previous works, in our experiments on nuScenes, all six input images are padded and resized from the original resolution of $1600 \times 900$ to $800\times480$. Similarly, for Argoverse 2, where the full input resolutions for the seven surround images are $2048\times1550$ for the front camera and $1550\times2048$ for the remaining six cameras, the inputs to our model are padded and rescaled to $640\times640$. For the BEV features $\mathbf{F}^{BEV}_{map}$ and $\mathbf{F}^{BEV}_{obj}$, we use sizes of $h_{obj} = w_{obj} = 150$ and $h_{map} = 200, w_{map} = 100$. In the tasks' decoders, we use $n_{obj} = 900$ queries for object detection and $p=20$ point queries for each of the $n'_{map}=50$ predicted map instances, resulting in a total of $n_{map}=1000$ map queries. On the nuScenes dataset, the model is trained for 24 epochs, on Argoverse 2 for 12 epochs. The epoch-dependent attention factor $\alpha$ is $0$ for the first 13 epochs of training on nuScenes and then linearly increased to a value of $0.14$ until the 16th epoch. On Argoverse 2, the same increase takes place from epochs 7 to 10. To scale the object detection loss to roughly the same magnitude as the map prediction loss, we set the scaling factor $\lambda$ to 3.0. All our experiments use a ResNet50\cite{he2016residual} backbone. Differing from Mask2Map \cite{mask2map} which was trained on 4 GPUs with a batch size of 4 each, all our training routines are performed on a single A100 GPU with a batch size of 4 using an AdamW optimizer with a weight decay of 0.01. This includes our reproduction of the Mask2Map baseline. The initial learning rate is set to $6\cdot10^{-4}$ for the training on nuScenes and $6\cdot10^{-5}$ for Argoverse 2 with cosine annealing in both cases. 

\section{Results}

\subsection{Results on nuScenes} 

In \Cref{nuscenes_results_table}, we present the performance metrics of our model ObjectGuidedMap in comparison to the most influential prior works \cite{maptr, mapvr, pivotnet, bemapnet, maptrv2, mask2map} evaluated on the nuScenes \cite{nuscenes_paper} validation set.

For transparency and comparability with prior work, we report both the originally published Mask2Map $mAP$ score of $0.716$ and the result obtained in our reproduction shown in \Cref{nuscenes_results_table}. Our improvement claim, however, is based on the controlled comparison between our reproduced Mask2Map baseline and ObjectGuidedMap. Both models use identical data preprocessing, optimization settings, training schedules, and evaluation code for the map prediction task. This comparison therefore assesses the effect of the proposed object-aware multitask extension within our experimental setup, while the published Mask2Map score serves as a literature reference for contextualizing absolute performance.

Our model achieves a $mAP$ of 0.661, outperforming the Mask2Map baseline by $4.99\%$ while maintaining significantly more stable performance over multiple training runs, as demonstrated by the much lower standard deviation of $0.008$ compared to $0.021$ for the baseline model. A similar pattern can also be observed for all map classes individually: ObjectGuidedMap outperforms the next best model for each class and achieves much more repeatable results across different training runs, with the improved stability for the curb class being particularly pronounced. 
\\
For the object detection task in our multitask model, we report the nuScenes Detection Score (NDS) \cite{nuscenes_paper}. Since the focus of this work is on the map prediction task, we did not perform any major optimization of the object prediction module. The achieved score is roughly equal to the performance of the comparable single task object detection model that pairs the BEVFormer transformer with the Mask2Map backbone and achieved an NDS of 0.304. This demonstrates that our method's improved map prediction does not come at the expense of deteriorated object detection performance, \ie, there is no negative transfer between the tasks. We leave the optimization of the object detection task as future work here. For an improved, more reliable object detection branch, we would expect an even larger impact on the map prediction task. 
\\
With the additional object detection branch, the size of our model is 77.9M parameters while Mask2Map uses 60.4M parameters. Table 9 in the supplementary material shows a more detailed size comparison between the two models. The number of floating point operations (FLOPs) required for one inference step is 885.15 GFLOPs for our ObjectGuidedMap model and 727.36 GFLOPs for Mask2Map. We argue that this moderate increase of 22\% in FLOPs can already be justified by the improved map prediction performance alone. However, any autonomous driving stack requires object detection anyway which is already covered by our model but still needs to be run separately alongside Mask2Map. With some additional finetuning of the object detection branch, this positions ObjectGuidedMap very competitively in terms of model size, computational cost and map prediction quality.

\begin{table*}[h!] 
\centering 
\caption{Results on the Argoverse 2 validation set. Relative improvements are shown in parentheses.} 
\resizebox{0.8\linewidth}{!}{ 
\begin{tabular}{l|C{2.4cm}|C{2.4cm}|C{2.4cm}|C{2.4cm}} 
\hline 
& \begin{tabular}{@{}c@{}} Lane Line\\ AP $\uparrow$ \end{tabular} & \begin{tabular}{@{}c@{}} Crosswalk\\ AP $\uparrow$ \end{tabular} & \begin{tabular}{@{}c@{}} Curb\\ AP $\uparrow$ \end{tabular} & mAP $\uparrow$ \\ 
\hline 
Ours w/o Cross-Task-Attn, DLF & 0.5082 & 0.3709 & 0.5405  & 0.4732  \\ 
Ours w/o DLF & 0.5289 (+4.07\%) & 0.3794 (+2.29\%) & 0.5679 (+5.07\%) & 0.4921 (+3.99\%) \\ 
ObjectGuidedMap (Ours) & \textbf{0.5339} (+5.06\%) & \textbf{0.4123}(+11.16\%) & \textbf{0.5808} (+7.46\%) & \textbf{0.5090} (+7.57\%) \\ 
\hline \end{tabular}
} 
\label{AV2_results_table} \end{table*}

\subsection{Cross-Dataset Ablation on Argoverse 2} To demonstrate the transferability of our method's impact across datasets, we show the results of our model on the Argoverse 2 dataset \cite{av2_paper} in \Cref{AV2_results_table}. We compare our complete ObjectGuidedMap model to a version that omits Deformable Latent Fusion (DLF) between the tasks and another that does not use any information from the object detection branch. All setups compared here share a common backbone for both tasks. These results demonstrate that our method is similarly effective on Argoverse 2 as it is on nuScenes, achieving a notable 11.16\% improvement for the crosswalk class and a 7.57\% improvement for the overall mAP. 

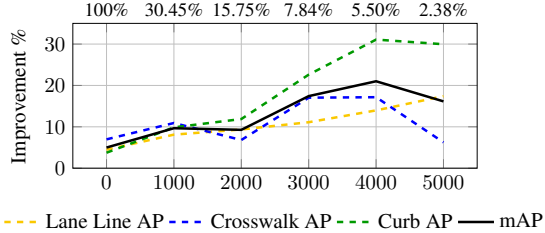
\begin{figure}
    \centering
    \resizebox{0.9\columnwidth}{!}{
    \begin{tikzpicture}
    \begin{axis}[
        width=\columnwidth,
        height=4cm,
        ymin=0, ymax=35,
        ylabel={Improvement \%},
        xtick={1,2,3,4,5,6},
        xticklabels={$0$, $1000$, $2000$, $3000$, $4000$, $5000$},
        legend style={
            at={(0.5,-0.25)},
            anchor=north,
            legend columns=4, 
            draw=none
        },
        grid=both, 
        clip=false
    ]
    
    \addplot[color={rgb,255:red,250; green,200; blue,0}, very thick, dashed] coordinates {
        (1,4.47) (2,8.11) (3,9.36) (4,11.11) (5,13.97) (6,17.40)
    };
    \addlegendentry{Lane Line AP}
    
    \addplot[color=blue, very thick, dashed] coordinates {
        (1,6.97) (2,10.92) (3,6.81) (4,17.04) (5,17.16) (6,6.25)
    };
    \addlegendentry{Crosswalk AP}
    
    \addplot[color=green!60!black, very thick, dashed] coordinates {
        (1,3.74) (2,9.87) (3,11.88) (4,22.61) (5,31.08) (6,29.96)
    };
    \addlegendentry{Curb AP}
    
    \addplot[color=black, very thick] coordinates {
        (1,4.99) (2,9.66) (3,9.27) (4,17.42) (5,21.01) (6,16.15)
    };
    \addlegendentry{mAP}

    \addplot[draw=none] coordinates {
        (1,35) (2,35) (3,35) (4,35) (5,35) (6,35)
    };
    \node at (axis cs:1,35) [anchor=south, font=\small] {100\%};
    \node at (axis cs:2,35) [anchor=south, font=\small] {30.45\%};
    \node at (axis cs:3,35) [anchor=south, font=\small] {15.75\%};
    \node at (axis cs:4,35) [anchor=south, font=\small] {7.84\%};
    \node at (axis cs:5,35) [anchor=south, font=\small] {5.50\%};
    \node at (axis cs:6,35) [anchor=south, font=\small] {2.38\%};
    
    \end{axis}
    \end{tikzpicture}
    }
    \caption{Improvement percentages on the nuScenes validation set relative to the Mask2Map baseline model for increasing amounts of occlusion quantified by the maximum number of LiDAR points for any object in a scene. The percentages at the top of the graph indicate the fraction of samples included in each occlusion level.}
    \label{occlusion_graph}
\end{figure}

\subsection{Results for Challenging Scenes} One of our method's primary goals is to improve map prediction in low-visibility and highly occluded environments. To quantify this improvement, we evaluate our model across scenes with varying degrees of occlusion. As a proxy for occlusion severity, we filter the nuScenes validation frames based on the maximum number of LiDAR points belonging to a single object in a scene, since the focus here is on occlusions caused by single large objects. The number of pixels occluded by an object is roughly proportional to the number of LiDAR points associated with it. \Cref{occlusion_graph} illustrates our relative performance improvement over the Mask2Map baseline as the occlusion threshold $N$ increases from 0 to 5,000 points. The results clearly show that our method performs well under heavy occlusion. The performance gap over the baseline increases significantly in scenes containing large occluding objects ($N > 3000$). However, for scenes with extreme occlusion levels ($N > 5000$), the relative improvement starts to deteriorate. We explain this with a lack of visual information. When a massive object blocks the view of the cameras, it becomes impossible to infer the underlying map geometry regardless of the model architecture. In Figure 8 in the supplementary material, we show such a failure case. 
We restrict the comparison in high-occlusion and poor-visibility scenarios to Mask2Map, as prior vectorized map prediction methods largely follow similar formulations that are crucially all single-task. There is little reason to assume that these models would perform significantly better in these difficult scenarios than Mask2Map.

\begin{table*}[t] \centering \caption{Improvement relative to the Mask2Map baseline on the full nuScenes validation set and on low-occlusion ($<100$ LiDAR points in total) and high-occlusion scenes (at least one object with $>4000$ LiDAR points). The percentages associated with the three subsets denote the fraction of validation samples included in each subset.} \resizebox{0.75\linewidth}{!}{\begin{tabular}{l|l||C{2.7cm}|C{2.7cm}|C{2.7cm}|C{2.7cm}} \hline & & \begin{tabular}{@{}c@{}} Lane Line\\ AP $\uparrow$ \end{tabular} & \begin{tabular}{@{}c@{}} Crosswalk\\ AP $\uparrow$ \end{tabular} & \begin{tabular}{@{}c@{}} Curb\\ AP $\uparrow$ \end{tabular} & mAP $\uparrow$ \\ \hline \multirow{2}{*}{Full Dataset (100\%)} & Mask2Map \cite{mask2map} & 0.6273 & 0.5864 & 0.6737 & 0.6291 \\ & ObjectGuidedMap & \textbf{0.6553 (+4.47\%)} & \textbf{0.6273 (+6.97\%)} & \textbf{0.6989 (+3.74\%)} & \textbf{0.6605 (+4.99\%)} \\ \hline \multirow{2}{*}{Low Occlusion (9.57\%)} & Mask2Map \cite{mask2map} & 0.5324 & \textbf{0.3614} & 0.6003 & 0.4980 \\ & ObjectGuidedMap & \textbf{0.5403 (+1.48\%)} & 0.3549 (-1.80\%) & \textbf{0.6375 (+6.20\%)} & \textbf{0.5109 (+2.59\%)} \\ \hline \multirow{2}{*}{High Occlusion (5.50\%)} & Mask2Map \cite{mask2map} & 0.3142 & 0.4918 & 0.4090 & 0.4050 \\ & ObjectGuidedMap & \textbf{0.3581 (+13.97\%)} & \textbf{0.5762 (+17.16\%)} & \textbf{0.5361 (+31.08\%)} & \textbf{0.4901 (+21.01\%)} \\ \hline \end{tabular}} \label{low_vs_high_occlusion_results_table} \end{table*}

\Cref{low_vs_high_occlusion_results_table} shows the performance metrics and improvement percentages for the full nuScenes dataset, low-occlusion, and high-occlusion scenes. 
Table 6 in the supplementary material additionally shows the metrics for the different matching thresholds. 
The evaluation on the full dataset and particularly on the low-occlusion scenes demonstrates that the improvement achieved in the high-occlusion cases does not come at the expense of performance on the other scenarios. It also shows, however, that in cases with very little occlusion, the impact of our method is minimal.
In the high-occlusion cases, we see a substantial increase of $0.085$ or $21\%$ in mAP and notably a particularly large improvement by 0.127 or $31\%$ in AP for the curb class. We hypothesize that in those scenes, curbs are the instances most likely to be too occluded for the baseline model to recognize them, while our model is still robust against the occlusion. This is usually due to vehicles parked curbside, a very common scenario in urban environments.

\begin{table*}[h!] \centering \caption{Improvement relative to the Mask2Map (M2M) baseline on the full nuScenes validation set, scenes with the ``rain'' or ``wet'' keywords, and scenes with the ``night'' keyword.} \resizebox{0.75\linewidth}{!}{\begin{tabular}{l|l||C{2.6cm}|C{2.6cm}|C{2.6cm}|C{2.6cm}} \hline & & \begin{tabular}{@{}c@{}} Lane Line\\ AP $\uparrow$ \end{tabular} & \begin{tabular}{@{}c@{}} Crosswalk\\ AP $\uparrow$ \end{tabular} & \begin{tabular}{@{}c@{}} Curb\\ AP $\uparrow$ \end{tabular} & mAP $\uparrow$ \\ \hline \multirow{2}{*}{} Full Dataset & M2M \cite{mask2map} & 0.6273 & 0.5864 & 0.6737 & 0.6291 \\ (100\%) & Ours & \textbf{0.6553} (+4.47\%) & \textbf{0.6273} (+6.97\%) & \textbf{0.6989} (+3.74\%) & \textbf{0.6605} (+4.99\%) \\ \hline \multirow{2}{*}{} Rain/Wet Only & M2M \cite{mask2map} & 0.5522 & 0.4265 & 0.5067 & 0.4951 \\ (18.08\%) & Ours & \textbf{0.6231} (+12.84\%) & \textbf{0.4868} (+14.14\%) & \textbf{0.5405} (+6.67\%) & \textbf{0.5501} (+11.11\%) \\ \hline \multirow{2}{*}{} Night Only & M2M \cite{mask2map} & 0.3767 & 0.1704 & 0.5777 & 0.3749 \\ (10.00\%) & Ours & \textbf{0.4259} (+13.06\%) & \textbf{0.3144} (+84.51\%) & \textbf{0.6183} (+7.03\%) & \textbf{0.4529} (+20.81\%) \\ \hline \end{tabular}} \label{rain_night_results_table} \end{table*}

\begin{figure}
    \centering
    \includegraphics[width=\columnwidth]{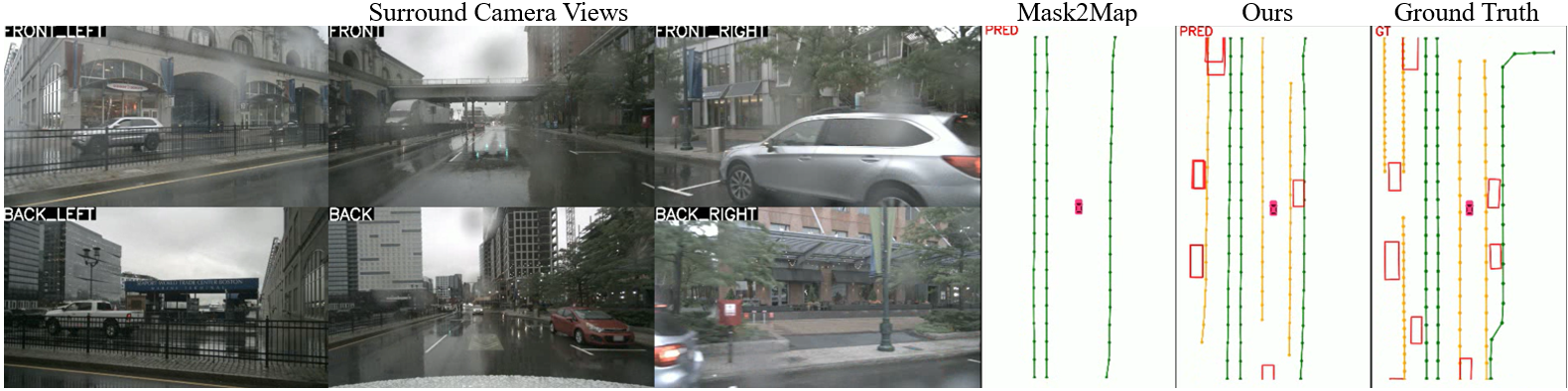}
    \caption{Comparison of results from our method to the Mask2Map \cite{mask2map} baseline: The wet road surface causes reflexions that render the baseline model unable to predict any lane lines while our model still remains robust, despite the poor visibility.}
    \label{improvement_example_picture_occ_rain}
\end{figure}

\Cref{rain_night_results_table} shows the results for rainy and night scenes, alongside the full dataset. For rainy scenarios, we observe larger improvements for lane lines and crosswalks while the improvement for curbs is not much larger than on the full dataset. This is also evident in \cref{improvement_example_picture_occ_rain}. Our interpretation of this result is that lane lines and crosswalks are two-dimensional road markings which makes them harder to recognize against reflections on a wet surface. Here, our model leverages valuable cues for the placement of these road markings by taking the positions of surrounding vehicles into account. Since curbs are not flat, they are not significantly harder to detect in these scenarios, explaining why there is little additional improvement compared to the full dataset for that class. 

Across night scenes, our model improves the crosswalk AP by $84.5\%$ over the Mask2Map baseline. We attribute this gain to the compact spatial extent of crosswalks, which often lie entirely at longer distances from the ego-vehicle and therefore provide few strong cues for full instance reconstruction. Under degraded long-range perception at night, this makes predicting crosswalks particularly challenging for the baseline model, resulting in low performance. Consequently, the substantial relative improvement is partly due to the especially weak baseline results in this setting. Fig. 7 in the supplementary material shows a comparison between the predictions from Mask2Map and ObjectGuidedMap for a complex night scene, illustrating the improved prediction quality of our model.

\begin{table*}[h!] \centering \caption{Ablation study of the main components of ObjectGuidedMap. SB denotes Shared Backbone, DLF denotes Deformable Latent Fusion, CTA denotes Cross-Task Attention, and AF denotes Attention Factors. Relative changes compared with the baseline are shown in parentheses.} \resizebox{0.75\linewidth}{!}{\begin{tabular}{c|c|c|c||C{2.6cm}|C{2.6cm}|C{2.6cm}|C{2.6cm}} \hline SB & DLF & CTA & AF & \begin{tabular}{@{}c@{}} Lane Line\\ AP $\uparrow$ \end{tabular} & \begin{tabular}{@{}c@{}} Crosswalk\\ AP $\uparrow$ \end{tabular} & \begin{tabular}{@{}c@{}} Curb\\ AP $\uparrow$ \end{tabular} & mAP $\uparrow$ \\ \hline - & - & - & - & 0.6273 & 0.5864 & 0.6737 & 0.6291 \\ \checkmark & - & - & - & 0.6134 (-2.21\%) & 0.5507 (-6.09\%) & 0.6675 (-0.92\%) & 0.6105 (-2.96\%) \\ \checkmark & - & \checkmark & \checkmark & 0.6511 (+3.80\%) & 0.6083 (+3.73\%) & 0.6833 (+1.42\%) & 0.6475 (+2.92\%) \\ \checkmark & \checkmark & \checkmark & - & 0.6552 (+4.46\%) & 0.6190 (+5.56\%) & 0.6796 (+0.87\%) & 0.6513 (+3.53\%) \\ \checkmark & \checkmark & - & \checkmark & 0.6535 (+4.18\%) & 0.6202 (+5.76\%) & 0.6823 (+1.27\%) & 0.6520 (+3.64\%) \\ \checkmark & \checkmark & \checkmark & \checkmark & \textbf{0.6553 (+4.47\%)} & \textbf{0.6273 (+6.97\%)} & \textbf{0.6989 (+3.74\%)} & \textbf{0.6605 (+4.99\%)} \\ \hline \end{tabular}} \label{ablation_table} \end{table*}

\subsection{Ablation Study} In \Cref{ablation_table}, we show the results of our ablation study to demonstrate the impact of the main features of ObjectGuidedMap. We evaluate the influence of sharing the backbone between the tasks, performing Deformable Latent Fusion (DLF) between the tasks' BEV feature, computing Cross-Task-Attention between the task query sets in the decoder, and whether epoch-dependent attention factors were used instead of a constant attention factor of $0.1$. All results are reported on the nuScenes validation set. The ablation study shows that only sharing the backbone leads to diminished performance compared to the single-task baseline due to its representational power now having to be shared between two tasks. Adding either of our attention modules to the model, however, more than compensates for that deterioration and already leads to a model that outperforms the baseline by a considerable margin. Similarly, using both attention modules, along with constant attention factors during training also results in stronger performance relative to the base Mask2Map model. Finally, using all features yields the largest improvement of $4.99\%$. This study demonstrates that each main idea of our method contributes a significant part to the model's overall performance.

\section{Conclusion}

In this work, we presented an object-aware multitask framework for vectorized BEV map prediction that jointly models object detection and map reconstruction. By integrating a detection branch and cross-attention modules for task interaction, the model leverages object-level cues to improve map prediction under occlusions and poor visibility. Extensive experiments demonstrate significant gains in challenging conditions while maintaining strong performance in favorable scenarios.
A current limitation of our approach is that it does not yet exploit temporal cues. 
Future work will focus on incorporating temporal information alongside cross-task information in unified attention modules. 
Overall, our results highlight that explicit relational conditioning is beneficial for structured scene reconstruction under partial observability. 

{
    \small
    \bibliographystyle{ieeenat_fullname}
    \bibliography{main}
}

\newpage
\section*{Supplementary Material}

In \Cref{full_results_table_occlusion}, we show the full comparison of the metrics achieved with the \\Mask2Map \cite{mask2map} baseline and our ObjectGuidedMap for several subsets of the nuScenes \cite{nuscenes_paper} validation set with different levels of occlusion. Notably, apart from the low-occlusion cases with fewer than 100 LiDAR points in total where the two models perform similarly well, our method significantly outperforms the baseline for every other subset of the validation data, every map instance class, and every matching threshold. 

\begin{table*}[h!]
    \centering
    \caption{Complete results from the Mask2Map baseline and our ObjectGuidedMap on the full nuScenes validation set and for different levels of occlusion quantified by the maximum number of LiDAR points for one object in the scene (* <100 points refers to the total number of points in a scene). In addition to the metrics in the main paper, for each class we report the AP for three different matching thresholds: 0.5m, 1.0m, and 1.5m.}
    \resizebox{\textwidth}{!}{
    \begin{tabular}{l|l||C{1.5cm}|C{1.5cm}|C{1.5cm}|C{1.5cm}||C{1.5cm}|C{1.5cm}|C{1.5cm}|C{1.5cm}||C{1.5cm}|C{1.5cm}|C{1.5cm}|C{1.5cm}||C{1.5cm}}
    \hline
    & & \begin{tabular}{@{}c@{}}Lane Line\\AP$_{0.5} \uparrow$\end{tabular} & \begin{tabular}{@{}c@{}}Lane Line\\AP$_{1.0} \uparrow$\end{tabular} & \begin{tabular}{@{}c@{}}Lane Line\\AP$_{1.5} \uparrow$\end{tabular} & \begin{tabular}{@{}c@{}}Lane Line\\AP $\uparrow$\end{tabular} & \begin{tabular}{@{}c@{}}Crosswalk\\AP$_{0.5} \uparrow$\end{tabular} & \begin{tabular}{@{}c@{}}Crosswalk\\AP$_{1.0} \uparrow$\end{tabular} & \begin{tabular}{@{}c@{}}Crosswalk\\AP$_{1.5} \uparrow$\end{tabular} & \begin{tabular}{@{}c@{}}Crosswalk\\AP $\uparrow$\end{tabular} & \begin{tabular}{@{}c@{}}Curb\\AP$_{0.5} \uparrow$\end{tabular} & \begin{tabular}{@{}c@{}}Curb\\AP$_{1.0} \uparrow$\end{tabular} & \begin{tabular}{@{}c@{}}Curb\\AP$_{1.5} \uparrow$\end{tabular} & \begin{tabular}{@{}c@{}}Curb\\AP $\uparrow$\end{tabular}  & mAP $\uparrow$\\
    \hline
    \multirow{2}{*}{Full Dataset} & Mask2Map \cite{mask2map} &0.503 &0.658 &0.721 & 0.627 &0.392 &0.633 &0.734 & 0.586 &0.493 &0.724 &0.805 & 0.674 & 0.629  \\
    & ObjectGuidedMap &\textbf{0.545}& \textbf{0.683}& \textbf{0.738}& \textbf{0.655}  &\textbf{0.435}& \textbf{0.679}& \textbf{0.768}&\textbf{0.627} &\textbf{0.521}& \textbf{0.752}& \textbf{0.823}&\textbf{0.699}&\textbf{0.661} \\
    \hline
    \multirow{2}{*}{$<100$ Points*} & Mask2Map \cite{mask2map} 
    & 0.403 & 0.560 & \textbf{0.634} & 0.532 
    & \textbf{0.192} & \textbf{0.382} & 0.510 & \textbf{0.361}  
    & 0.409 & 0.644 & 0.748 & 0.600  & 0.498 \\
    & ObjectGuidedMap 
    & \textbf{0.423} & \textbf{0.564} & \textbf{0.634} & \textbf{0.540} 
    & 0.179 & 0.363 & \textbf{0.522} & 0.355 
    & \textbf{0.470} & \textbf{0.672} & \textbf{0.771} & \textbf{0.638}  & \textbf{0.511}  \\
    \hline
    \multirow{2}{*}{$>1000$ Points} & Mask2Map \cite{mask2map} 
    & 0.464& 0.606& 0.676& 0.582 
    & 0.434& 0.664& 0.739& 0.613 
    & 0.466& 0.700& 0.785& 0.650 & 0.615\\
    & ObjectGuidedMap   
    & \textbf{0.522}& \textbf{0.657} &\textbf{0.708}& \textbf{0.629} 
    & \textbf{0.505}& \textbf{0.729}& \textbf{0.804}& \textbf{0.680} 
    & \textbf{0.546}& \textbf{0.766}& \textbf{0.832}& \textbf{0.715} & \textbf{0.674}\\
    \hline
    \multirow{2}{*}{$>2000$ Points} & Mask2Map \cite{mask2map} 
    & 0.408& 0.536& 0.610& 0.518 
    & 0.562& 0.711& 0.767& 0.680 
    & 0.452& 0.666& 0.756& 0.625 & 0.608\\
    & ObjectGuidedMap 
    & \textbf{0.455}& \textbf{0.596}& \textbf{0.648}& \textbf{0.566} 
    & \textbf{0.609}& \textbf{0.755}& \textbf{0.814}& \textbf{0.726} 
    & \textbf{0.526}& \textbf{0.753}& \textbf{0.818}& \textbf{0.699} & \textbf{0.664}\\
    \hline
    \multirow{2}{*}{$>3000$ Points} & Mask2Map \cite{mask2map} 
    & 0.277& 0.370& 0.455& 0.367 
    & 0.321& 0.566& 0.626& 0.504 
    & 0.282& 0.531& 0.639& 0.484 & 0.452\\
    & ObjectGuidedMap 
    & \textbf{0.304}& \textbf{0.435}& \textbf{0.486}& \textbf{0.408} 
    & \textbf{0.397}& \textbf{0.650}& \textbf{0.723}& \textbf{0.590} 
    & \textbf{0.398}& \textbf{0.658}& \textbf{0.725}& \textbf{0.594} & \textbf{0.531}\\
    \hline
    \multirow{2}{*}{$>4000$ Points} & Mask2Map \cite{mask2map} 
    & 0.232& 0.315& 0.396& 0.314  
    & 0.308& 0.560& 0.607& 0.492  
    & 0.204 & 0.450 & 0.573& 0.409  & 0.405  \\
    & ObjectGuidedMap 
    & \textbf{0.258}& \textbf{0.385}& \textbf{0.431}& \textbf{0.358}  
    & \textbf{0.402}& \textbf{0.634}& \textbf{0.692}& \textbf{0.576}  
    & \textbf{0.325}& \textbf{0.600}& \textbf{0.683}& \textbf{0.536}  & \textbf{0.490}  \\
    \hline
    \multirow{2}{*}{$>5000$ Points} & Mask2Map \cite{mask2map} 
    & 0.448& 0.595& 0.658& 0.567 
    & 0.420& 0.678& 0.789& 0.629 
    & 0.293& 0.491& 0.749& 0.511 & 0.569 \\
    & ObjectGuidedMap  
    & \textbf{0.565}& \textbf{0.693}& \textbf{0.740}& \textbf{0.666} 
    & \textbf{0.471}& \textbf{0.711}& \textbf{0.824}& \textbf{0.668} 
    & \textbf{0.441}& \textbf{0.692}& \textbf{0.812}& \textbf{0.649} & \textbf{0.661}\\
    \hline
    \end{tabular}
    }
    \label{full_results_table_occlusion}
\end{table*}

\Cref{full_results_table_night_rain} shows the full comparison of the metrics achieved with the Mask2Map \cite{mask2map} baseline and our ObjectGuidedMap for the full nuScenes \cite{nuscenes_paper} validation set, and subsets filtered for the "rain"/"wet" and "night" keywords. Our method outperforms the baseline model by a significant margin for every map instance class, every subset of the data, and every matching threshold. As mentioned in the main paper, of particular note are the improvements for the crosswalk class at night: With an improvement of 268\% for the 0.5m matching threshold, our model achieves the highest observed impact here.

\begin{table*}[h!]
    \centering
    \caption{Complete results from the Mask2Map baseline and our ObjectGuidedMap on the full nuScenes validation set and filtered for scenes with the "rain"/"wet" and "night" keyword. In addition to the metrics in the main paper, for each class we report the AP for three different matching thresholds: 0.5m, 1.0m, and 1.5m.}
    \resizebox{\textwidth}{!}{
    \begin{tabular}{l|l||C{1.5cm}|C{1.5cm}|C{1.5cm}|C{1.5cm}||C{1.5cm}|C{1.5cm}|C{1.5cm}|C{1.5cm}||C{1.5cm}|C{1.5cm}|C{1.5cm}|C{1.5cm}||C{1.5cm}}
    \hline
    & & \begin{tabular}{@{}c@{}}Lane Line\\AP$_{0.5} \uparrow$\end{tabular} & \begin{tabular}{@{}c@{}}Lane Line\\AP$_{1.0} \uparrow$\end{tabular} & \begin{tabular}{@{}c@{}}Lane Line\\AP$_{1.5} \uparrow$\end{tabular} & \begin{tabular}{@{}c@{}}Lane Line\\AP $\uparrow$\end{tabular} & \begin{tabular}{@{}c@{}}Crosswalk\\AP$_{0.5} \uparrow$\end{tabular} & \begin{tabular}{@{}c@{}}Crosswalk\\AP$_{1.0} \uparrow$\end{tabular} & \begin{tabular}{@{}c@{}}Crosswalk\\AP$_{1.5} \uparrow$\end{tabular} & \begin{tabular}{@{}c@{}}Crosswalk\\AP $\uparrow$\end{tabular} & \begin{tabular}{@{}c@{}}Curb\\AP$_{0.5} \uparrow$\end{tabular} & \begin{tabular}{@{}c@{}}Curb\\AP$_{1.0} \uparrow$\end{tabular} & \begin{tabular}{@{}c@{}}Curb\\AP$_{1.5} \uparrow$\end{tabular} & \begin{tabular}{@{}c@{}}Curb\\AP $\uparrow$\end{tabular}  & mAP $\uparrow$\\
    \hline
    \multirow{2}{*}{Full Dataset} & Mask2Map \cite{mask2map} &0.503 &0.658 &0.721 & 0.627 &0.392 &0.633 &0.734 & 0.586 &0.493 &0.724 &0.805 & 0.674 & 0.629  \\
    & ObjectGuidedMap &\textbf{0.545}& \textbf{0.683}& \textbf{0.738}& \textbf{0.655}  &\textbf{0.435}& \textbf{0.679}& \textbf{0.768}&\textbf{0.627} &\textbf{0.521}& \textbf{0.752}& \textbf{0.823}&\textbf{0.699}&\textbf{0.661} \\
    \hline
    \multirow{2}{*}{Rain/Wet Only} & Mask2Map \cite{mask2map} &0.419& 0.580& 0.657& 0.552 &0.193& 0.460&0.626& 0.427 &0.298& 0.557& 0.665& 0.507 & 0.495 \\
    & ObjectGuidedMap &\textbf{0.487}& \textbf{0.652}& \textbf{0.730}& \textbf{0.623} &\textbf{0.243}&\textbf{0.543}& \textbf{0.674}& \textbf{0.487} &\textbf{0.353}&\textbf{ 0.587}&\textbf{0.681}&\textbf{0.541}& \textbf{0.550}\\
    \hline
    \multirow{2}{*}{Night Only} & Mask2Map \cite{mask2map} &0.217& 0.403& 0.510& 0.377 &0.036 & 0.198& 0.278& 0.170 &0.351& 0.636& 0.746& 0.578 & 0.375  \\
    & ObjectGuidedMap  &\textbf{0.255}& \textbf{0.461}& \textbf{0.562}& \textbf{0.426} &\textbf{0.131} & \textbf{0.301}& \textbf{0.511}& \textbf{0.314} &\textbf{0.393}& \textbf{0.686}& \textbf{0.776}& \textbf{0.618} & \textbf{0.453} \\
    \hline
    \end{tabular}
    }
    \label{full_results_table_night_rain}
\end{table*}

In \Cref{full_av2_results_table}, we show the full results of our evaluation on the Argoverse 2 \cite{av2_paper} dataset. These results show that the addition of Cross-Task-Attention and Deformable Latent Fusion modules improves the metrics for every matching threshold and any map instance class. 

\begin{table*}[h!]
    \centering
    \caption{Complete results on the Argoverse2 validation set. In addition to the metrics in the main paper, for each class we report the AP for three different matching thresholds: 0.5m, 1.0m, and 1.5m.}
    \resizebox{\textwidth}{!}{
    \begin{tabular}{l||C{1.5cm}|C{1.5cm}|C{1.5cm}|C{1.5cm}||C{1.5cm}|C{1.5cm}|C{1.5cm}|C{1.5cm}||C{1.5cm}|C{1.5cm}|C{1.5cm}|C{1.5cm}||C{1.5cm}}
    \hline
    & \begin{tabular}{@{}c@{}}Lane Line\\AP$_{0.5} \uparrow$\end{tabular} & \begin{tabular}{@{}c@{}}Lane Line\\AP$_{1.0} \uparrow$\end{tabular} & \begin{tabular}{@{}c@{}}Lane Line\\AP$_{1.5} \uparrow$\end{tabular} & \begin{tabular}{@{}c@{}}Lane Line\\AP $\uparrow$\end{tabular} & \begin{tabular}{@{}c@{}}Crosswalk\\AP$_{0.5} \uparrow$\end{tabular} & \begin{tabular}{@{}c@{}}Crosswalk\\AP$_{1.0} \uparrow$\end{tabular} & \begin{tabular}{@{}c@{}}Crosswalk\\AP$_{1.5} \uparrow$\end{tabular} & \begin{tabular}{@{}c@{}}Crosswalk\\AP $\uparrow$\end{tabular} & \begin{tabular}{@{}c@{}}Curb\\AP$_{0.5} \uparrow$\end{tabular} & \begin{tabular}{@{}c@{}}Curb\\AP$_{1.0} \uparrow$\end{tabular} & \begin{tabular}{@{}c@{}}Curb\\AP$_{1.5} \uparrow$\end{tabular} & \begin{tabular}{@{}c@{}}Curb\\AP $\uparrow$\end{tabular}  & mAP $\uparrow$\\
    \hline
    ObjectGuidedMap w/o Cross-Task-Attn., DLF &0.377&0.533&0.614& 0.508 &0.056&0.420&0.637& 0.371 &0.260&0.604&0.759& 0.541 & 0.473 \\
    ObjectGuidedMap w/o DLF &0.398&\textbf{0.557}&0.632& 0.529 &0.069&0.422&0.648& 0.379 &0.294&0.634&0.777& 0.568 & 0.492 \\
    ObjectGuidedMap & \textbf{0.407} & \textbf{0.557} & \textbf{0.637} & \textbf{0.534} & \textbf{0.095} & \textbf{0.463} & \textbf{0.678} & \textbf{0.412} & \textbf{0.307} & \textbf{0.645} & \textbf{0.790} & \textbf{0.581} & \textbf{0.509} \\
    \hline
    \end{tabular}
    }
    \label{full_av2_results_table}
\end{table*}

\Cref{model_overview_table} details the size of the individual components of both the Mask2Map \cite{mask2map} baseline model and our ObjectGuidedMap. While our model is roughly 30\% larger in terms of trainable parameters, it also performs the additional task of object detection which would otherwise need to be handled by a fully separate model alongside Mask2Map. 

\begin{table*}[h!]
    \centering
    \caption{Size comparison between our ObjectGuidedMap and Mask2Map\cite{mask2map}. Cross-Task Attention subsumes the attention layers, feedforward networks and normalizations for all refinement layers in that module. FPN denotes a Feature Pyramid Network.}
    \resizebox{0.7\textwidth}{!}{
    \begin{tabular}
    {l|r||l|r}
    \hline
    \multicolumn{2}{c||}{\textbf{ObjectGuidedMap}} & \multicolumn{2}{c}{\textbf{Mask2Map}} \\
    \hline
    Image Backbone & 23.5M & Image Backbone & 23.5M\\
    Image Neck (FPN) & 1.1M & Image Neck (FPN) & 1.1M\\
    Map Prediction Branch & 35.8M &  Map Prediction Branch & 35.8M\\
    Object Detection Branch & 14.0M & - &-\\
    Deformable Latent Fusion & 0.2M & - &-\\
    Cross-Task Attention & 3.2M & - &-\\
    \hline
    Full Model & 77.9M & Full Model & 60.4M \\
    \hline
    \end{tabular}
    }
    \label{model_overview_table}
\end{table*}

\begin{figure*}
    \centering
    \includegraphics[width=\textwidth]{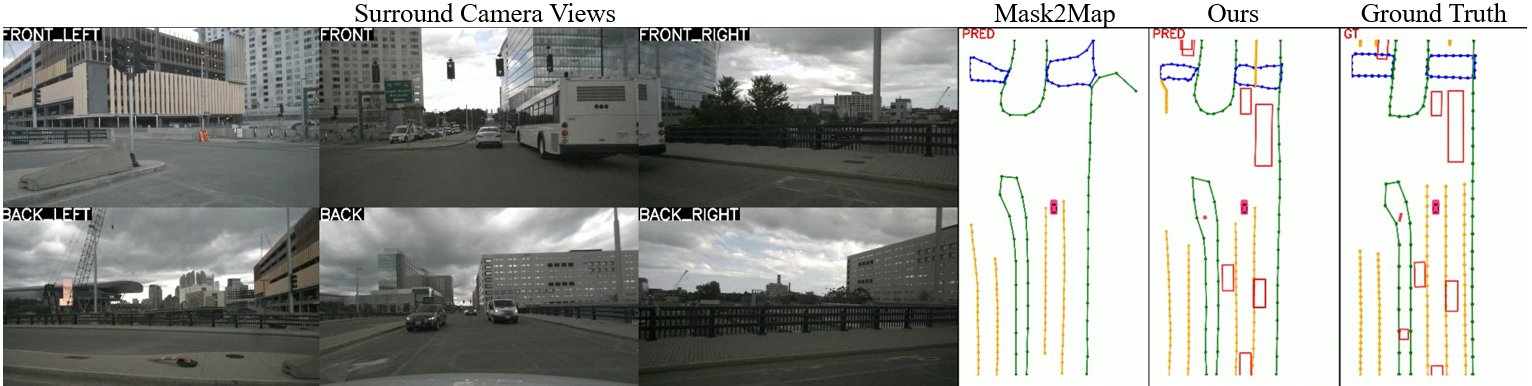}
    \caption{Comparison of results from our method to the Mask2Map \cite{mask2map} baseline for an example with a large occlusion}
    \label{improvement_example_picture_occ}
\end{figure*}

In \Cref{improvement_example_picture_occ}, we show an additional example of a large occlusion where our model outperforms the Mask2Map \cite{mask2map} baseline. The bus in the right lane in front of the ego-vehicle occludes a large part of the crosswalk ahead and of the curb on the right. Here, our method remains robust and produces a result that is much closer to the ground truth annotation compared to the baseline whose prediction is severely degraded in the area occluded by the bus. 

\begin{figure*}
    \centering
    \includegraphics[width=\textwidth]{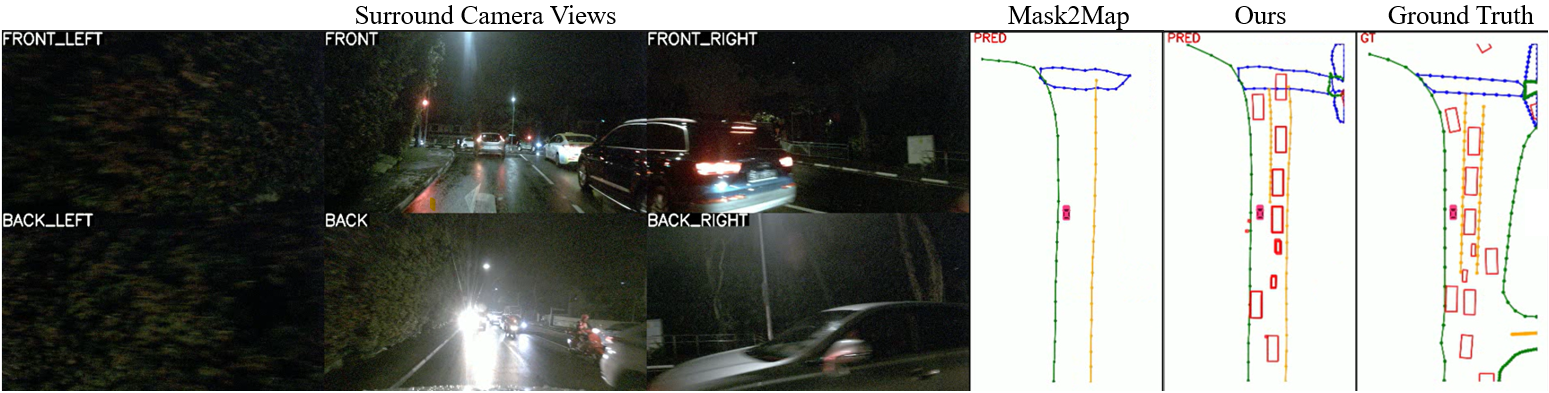}
    \caption{Comparison of results from our ObjectGuidedMap to the Mask2Map \cite{mask2map} baseline for a night scenario}
    \label{night_example}
\end{figure*}

\Cref{night_example} depicts the comparison of the predictions from our model and \\Mask2Map \cite{mask2map} for a night scene.
While both models struggle to accurately reconstruct the surrounding map at night, due to night scenes being severely underrepresented in the training data, ObjectGuidedMap still produces a more detailed map prediction compared to the baseline.

\begin{figure*}
    \centering
    \includegraphics[width=\textwidth]{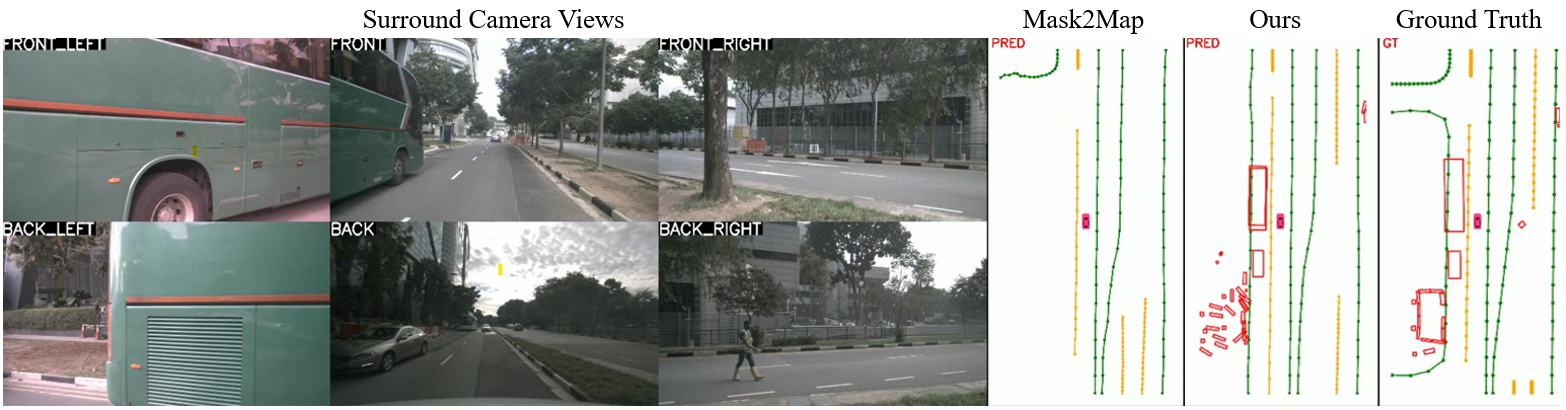}
    \caption{Failure case with extreme level of occlusion}
    \label{failure_case_extreme_occl}
\end{figure*}

\Cref{failure_case_extreme_occl} shows a failure case of our method: In this example, ObjectGuidedMap is unable to precisely reconstruct the curb to the left of the ego-vehicle due to the extreme occlusion caused by the parked bus.

\end{document}